# Sentiment Analysis of Twitter Data: A Survey of Techniques

Vishal A. Kharde
Department of Computer Engg,
Pune Institute of Computer Technology,Pune
University of Pune (India)

S.S. Sonawane
Department of Computer Engg,
Pune Institute of Computer Technology,Pune
University of Pune (India)

## ABSTRACT
With the advancement of web technology and its growth, there is a huge volume of data present in the web for internet users and a lot of data is generated too. Internet has become a platform for online learning, exchanging ideas and sharing opinions. Social networking sites like Twitter, Facebook, Google+ are rapidly gaining popularity as they allow people to share and express their views about topics, have discussion with different communities, or post messages across the world. There has been lot of work in the field of sentiment analysis of twitter data. This survey focuses mainly on sentiment analysis of twitter data which is helpful to analyze the information in the tweets where opinions are highly unstructured, heterogeneous and are either positive or negative, or neutral in some cases. In this paper, we provide a survey and a comparative analyses of existing techniques for opinion mining like machine learning and lexicon-based approaches, together with evaluation metrics. Using various machine learning algorithms like Naive Bayes, Max Entropy, and Support Vector Machine, we provide research on twitter data streams.We have also discussed general challenges and applications of Sentiment Analysis on Twitter.

## 1. INTRODUCTION
Nowadays, the age of Internet has changed the way people express their views, opinions. It is now mainly done through blog posts, online forums, product review websites, social media ,etc. Nowadays, millions of people are using social network sites like Facebook, Twitter, Google Plus, etc. to express their emotions, opinion and share views about their daily lives. Through the online communities, we get an interactive media where consumers inform and influence others through forums. Social media is generating a large volume of sentiment rich data in the form of tweets, status updates, blog posts, comments, reviews, etc. Moreover, social media provides an opportunity for businesses by giving a platform to connect with their customers for advertising. People mostly depend upon user generated content over online to a great extent for decision making. For e.g. if someone wants to buy a product or wants to use any service, then they firstly look up its reviews online, discuss about it on social media before taking a decision. The amount of content generated by users is too vast for a normal user to analyze. So there is a need to automate this, various sentiment analysis techniques are widely used.

Sentiment analysis (SA)tells user whether the information about the product is satisfactory or not before they buy it. Marketers and firms use this analysis data to understand about their products or services in such a way that it can be offered as per the user's requirements.

Textual Information retrieval techniques mainly focus on processing, searching or analyzing the factual data present. Facts have an objective component but,there are some other textual contents which express subjective characteristics. These contents are mainly opinions, sentiments, appraisals, attitudes, and emotions, which form the core of Sentiment Analysis (SA). It offers many challenging opportunities to develop new applications, mainly due to the huge growth of available information on online sources like blogs and social networks. For example, recommendations of items proposed by a recommendation system can be predicted by taking into account considerations such as positive or negative opinions about those items by making use of SA.

## 2. SENTIMENT ANALYSIS
Sentiment analysis can be defined as a process that automates mining of attitudes, opinions, views and emotions from text, speech, tweets and database sources through Natural Language Processing (NLP). Sentiment analysis involves classifying opinions in text into categories like "positive" or "negative" or "neutral". It's also referred as subjectivity analysis, opinion mining, and appraisal extraction.

The words opinion, sentiment, view and belief are used interchangeably but there are differences between them.

- *Opinion*: A conclusion open to dispute (because different experts have different opinions )
- *View:* subjective opinion
- *Belief:* deliberate acceptance and intellectual assent
- *Sentiment:* opinion representing one's feelings

An example for terminologies for Sentiment Analysis is as given below,

<SENTENCE> = The story of the movie was weak and boring

<OPINION HOLDER> =<author>

<OBJECT> = <movie>

<FEATURE> = <story>

<OPINION >= <weak><boring>

<POLARITY> = <negative>

Sentiment Analysis is a term that include many tasks such as sentiment extraction, sentiment classification, subjectivity classification, summarization of opinions or opinion spam detection, among others. It aims to analyze people's sentiments, , attitudes, opinions emotions, etc. towards elements such as, products, individuals, topics ,organizations, and services.

Mathematically we can represent an opinion as a quintuple (o, f, so, h, t), where

o = object;

f = feature of the object o;

so = orientation or polarity of the opinion on feature f of object o;

h = opinion holder;

t = time when the opinion is expressed.

*Object:* An entity which can be a, person, event, product, organization, or topic

*Feature:* An attribute (or a part) of the object with respect to which evaluation is made.

*Opinion orientation or polarity:* The orientation of an opinion on a feature f represent whether the opinion is positive, negative or neutral.

*Opinion holder*: The holder of an opinion is the person or organization or an entity that expresses the opinion.

In recent years a lot of work has been done in the field of "Sentiment Analysis on Twitter" by number of researchers. In its early stage it was intended for binary classification which assigns opinions or reviews to bipolar classes such as positive or negative only.

Pak and Paroubek(2010) [1] proposed a model to classify the tweets as objective, positive and negative. They created a twitter corpus by collecting tweets using Twitter API and automatically annotating those tweets using emoticons. Using that corpus, they developed a sentiment classifier based on the multinomial Naive Bayes method that uses features like N-gram and POS-tags. The training set they used was less efficient since it contains only tweets having emoticons.

Parikh and Movassate(2009) [2] implemented two models, a Naive Bayes bigram model and a Maximum Entropy model to classify tweets. They found that the Naive Bayes classifiers worked much better than the Maximum Entropy model.

Go and L.Huang (2009) [3] proposed a solution for sentiment analysis for twitter data by using distant supervision, in which their training data consisted of tweets with emoticons which served as noisy labels. They build models using Naive Bayes, MaxEnt and Support Vector Machines (SVM). Their feature space consisted of unigrams, bigrams and POS. They concluded that SVM outperformed other models and that unigram were more effective as features.

Barbosa et al.(2010) [4] designed a two phase automatic sentiment analysis method for classifying tweets. They classified tweets as objective or subjective and then in second phase, the subjective tweets were classified as positive or negative. The feature space used included retweets, hashtags, link, punctuation and exclamation marks in conjunction with features like prior polarity of words and POS.

Bifet and Frank(2010) [5] used Twitter streaming data provided by Firehouse API , which gave all messages from every user which are publicly available in real-time. They experimented multinomial naive Bayes, stochastic gradient descent, and the Hoeffding tree. They arrived at a conclusion that SGD-based model, when used with an appropriate learning rate was the better than the rest used.

Agarwal et al. (2011)[6] developed a 3-way model for classifying sentiment into positive, negative and neutral classes. They experimented with models such as: unigram model, a feature based model and a tree kernel based model. For tree kernel based model they represented tweets as a tree. The feature based model uses 100 features and the unigram model uses over 10,000 features. They arrived on a conclusion that features which combine prior polarity of words with their parts-of-speech(pos) tags are most important and plays a major role in the classification task. The tree kernel based model outperformed the other two models.

Davidov et al.,(2010) [7] proposed a approach to utilize Twitter user-defined hastags in tweets as a classification of sentiment type using punctuation, single words, n-grams and patterns as different feature types, which are then combined into a single feature vector for sentiment classification. They made use of K-Nearest Neighbor strategy to assign sentiment labels by constructing a feature vector for each example in the training and test set.

Po-Wei Liang et.al.(2014) [8] used Twitter API to collect twitter data. Their training data falls in three different categories (camera, movie , mobile). The data is labeled as positive, negative and non-opinions. Tweets containing opinions were filtered. Unigram Naive Bayes model was implemented and the Naive Bayes simplifying independence assumption was employed. They also eliminated useless features by using the Mutual Information and Chi square feature extraction method. Finally, the orientation of an tweet is predicted. i.e. positive or negative.

Pablo et. al. [9] presented variations of Naive Bayes classifiers for detecting polarity of English tweets. Two different variants of Naive Bayes classifiers were built namely Baseline (trained to classify tweets as positive, negative and neutral), and Binary (makes use of a polarity lexicon and classifies as positive and negative. Neutral tweets neglected). The features considered by classifiers were Lemmas (nouns, verbs, adjectives and adverbs), Polarity Lexicons, and Multiword from different sources and Valence Shifters.

Turney et al [11] used bag-of-words method for sentiment analysis in which the relationships between words was not at all considered and a document is represented as just a collection of words. To determine the sentiment for the whole document, sentiments of every word was determined and those values are united with some aggregation functions.

Kamps et al. [12] used the lexical database WordNet to determine the emotional content of a word along different dimensions. They developed a distance metric on WordNet and determined semantic polarity of adjectives.

Xia et al. [13] used an ensemble framework for Sentiment Classification which is obtained by combining various feature sets and classification techniques. In thier work, they used two types of feature sets (Part-of-speech information and Word-relations) and three base classifiers (Naive Bayes, Maximum Entropy and Support Vector Machines) . They applied ensemble approaches like fixed combination, weighted combination and Meta-classifier combination for sentiment classification and obtained better accuracy.

Luoet. al. [14] highlighted the challenges and an efficient techniques to mine opinions from Twitter tweets. Spam and wildly varying language makes opinion retrieval within Twitter challenging task.

A General model for sentiment analysis is as follows,

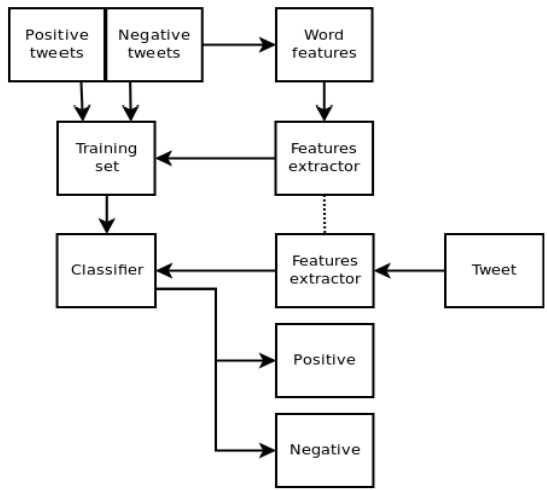

**Fig.1. Sentiment Analysis Architecture**

Following are the phases required for sentiment analysis of twitter data,

## 2.1 Pre-processing of the datasets
A tweet contains a lot of opinions about the data which are expressed in different ways by different users .The twitter dataset used in this survey work is already labeled into two classes viz. negative and positive polarity and thus the sentiment analysis of the data becomes easy to observe the effect of various features. The raw data having polarity is highly susceptible to inconsistency and redundancy. Preprocessing of tweet include following points,

- Remove all URLs (e.g. www.xyz.com), hash tags (e.g. #topic), targets (@username)
- Correct the spellings; sequence of repeated characters is to be handled
- Replace all the emoticons with their sentiment.
- Remove all punctuations ,symbols, numbers
- Remove Stop Words
- Expand Acronyms(we can use a acronym dictionary)
- Remove Non-English Tweets

**Table 1. Publicly Available Datasets For Twitter**

| HASH | Tweets | http://demeter.inf.ed.ac.uk | 31,861 Pos tweets 64,850 Neg tweets, 125,859 Neu tweets |
|---|---|---|---|
| EMOT | Tweets and Emoticons | http://twittersentiment.appspot.com | 230,811 Pos& 150,570 Neg tweets |
| ISIEVE | Tweets | www.i-sieve.com | 1,520 Pos tweets,200 Neg tweets, 2,295 Neu tweets |
| Columbia univ.dataset | Tweets | Email: apoorv@cs.columbia.edu | 11,875 tweets |
| Patient dataset | Opinions | http://patientopinion.org.uk | 2000 patient opinions |
| Sample | Tweets | http://goo.gl/UQvdx | 667 tweets |
| Stanford dataset | Movie Reviews | http://ai.stanford.edu/~amaas/data/sentiment/ | 50000 movie reviews |
| Stanford | Tweets | http://cs.stanford.edu/people/alecmgo/trainingandtestdata.zip | 4 million tweets categorized as positive and negative |
| Spam dataset | Spam Reviews | http://myleott.com/op_spam | 400 deceptive and 400 truthful reviews in positive and negative category. |
| Soe dataset | Sarcasm and nasty reviews | http://nlds.soe.ucsc.edu/iac | 1,000 discussions, ~390,000 posts, and some ~ 73,000,000 words |

## 2.2 Feature Extraction
The preprocessed dataset has many distinctive properties. In the feature extraction method, we extract the aspects from the processed dataset. Later this aspect are used to compute the positive and negative polarity in a sentence which is useful for determining the opinion of the individuals using models like unigram, bigram [18].

Machine learning techniques require representing the key features of text or documents for processing. These key features are c o n s i d e r e d as feature vectors which are used for the classification task..Some examples features that have been reported in literature are:

*1. Words And Their Frequencies:*
Unigrams, bigrams and n-gram models with their frequency counts are considered as features. There has been more research on using word presence rather than frequencies to better describe this feature. Panget al. [23] showed better results by using presence instead of frequencies.

*2. Parts Of Speech Tags*
Parts of speech like adjectives, adverbs and somegroups of verbs and nouns are good indicators of subjectivity and sentiment. We can generate syntactic dependency patterns by parsing or dependency trees.

*3. Opinion Words And Phrases*
Apart from specific words, some phrases and idioms which convey sentiments can be used as features.
e.g. cost someone an arm and leg.

*4. Position Of Terms*
The position of a term with in a text can affect on how much the term makes difference in overall sentiment of the text.

*5. Negation*
Negation is an important but difficult feature to interpret. The presence of a negation usually changes the polarity of the opinion..

e.g., I am not happy.

*6. Syntax*

Syntactic patterns like collocations are used as features to learn subjectivity patterns by many of the researchers.

## 2.3 Training

Supervised learning is an important technique for solving classification problems. Training the classifier makes it easier for future predictions for unknown data.

## 2.4 Classification

*2.4.1 Naive Bayes:*

It is a probabilistic classifier and can learn the pattern of examining a set of documents that has been categorized [9]. It compares the contents with the list of words to classify the documents to their right category or class. Let d be the tweet and c* be a class that is assigned to d, where

$$C^* = \arg mac_c P_{NB}(c|d)$$

$$P_{NB}(c|d) = \frac{(P(c))\sum_{i=1}^{m} p(f|c)^{n_i(d)}}{P(d)}$$

From the above equation, 'f' is a 'feature', count of feature (fi) is denoted with $n_i(d)$ and is present in d which represents a tweet. Here, m denotes no. of features.

Parameters P(c) and P(f|c) are computed through maximum likelihood estimates, and smoothing is utilized for unseen features. To train and classify using Naïve Bayes Machine Learning technique, we can use the Python NLTK library.

*2.4.2 Maximum Entropy*

In Maximum Entropy Classifier, no assumptions are taken regarding the relationship in between the features extracted from dataset. This classifier always tries to maximize the entropy of the system by estimating the conditional distribution of the class label.

Maximum entropy even handles overlap feature and is same as logistic regression method which finds the distribution over classes. The conditional distribution is defined as MaxEnt makes no independence assumptions for its features, unlike Naive Bayes.

The model is represented by the following:

$$P_{ME}(c|d,\lambda) = \frac{\exp[\sum_i \lambda_i f_i(c,d)]}{\sum_c \exp[\sum_i \lambda_i f_i(c,d)]}$$

Where c is the class,d is the tweet and $\lambda_i$ is the weight vector.The weight vectors decide the importance of a feature in classification.

*2.4.3 Support Vector Machine:*

Support vector machine analyzes the data, define the decision boundaries and uses the kernels for computation which are performed in input space[15]. The input data are two sets of vectors of size m each. Then every data which represented as a vector is classified into a class. Nextly we find a margin between the two classes that is far from any document. The distance defines the margin of the classifier, maximizing the margin reduces indecisive decisions. SVM also supports classification and regression which are useful for statistical learning theory and it also helps recognizing the factors precisely, that needs to be taken into account, to understand it successfully.

## 3. APPROACHES FOR SENTIMENT ANALYSIS

There are mainly two techniques for sentiment analysis for the twitter data:

## 3.1 Machine Learning Approaches

Machine learning based approach uses classification technique to classify text into classes. There are mainly two types of machine learning techniques

*3.1.1. Unsupervised learning:*

It does not consist of a category and they do not provide with the correct targets at all and therefore rely on clustering.

*3.1.2. Supervised learning:*

It is based on labeled dataset and thus the labels are provided to the model during the process. These labeled dataset are trained to get meaningful outputs when encountered during decision-making.

The success of both this learning methods is mainly depends on the selection and extraction of the specific set of features used to detect sentiment.

The machine learning approach applicable to sentiment analysis mainly belongs to supervised classification. In a machine learning techniques, two sets of data are needed:

1. Training Set
2. Test Set.

A number of machine learning techniques have been formulated to classify the tweets into classes. Machine learning techniques like Naive Bayes (NB), maximum entropy (ME), and support vector machines (SVM) have achieved great success in sentiment analysis.

Machine learning starts with collecting training dataset. Nextly we train a classifier on the training data. Once a supervised classification technique is selected, an important decision to make is to select feature. They can tell us how documents are represented.

The most commonly used features in sentiment classification are

- Term presence and their frequency
- Part of speech information
- Negations
- Opinion words and phrases

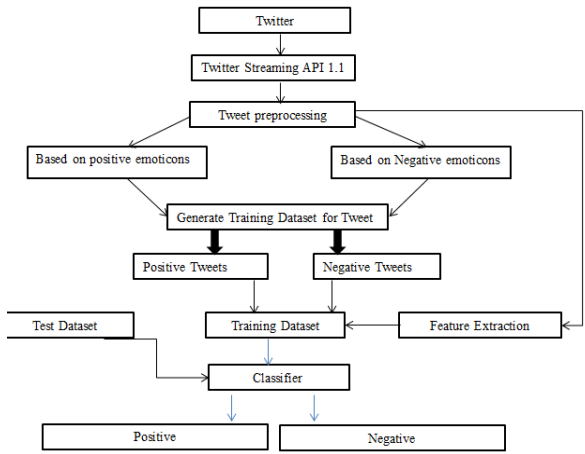

**Fig.2 Sentiment Classification Based On Emoticons**

With respect to supervised techniques, support vector machines (SVM), Naive Bayes, Maximum Entropy are some of the most common techniques used.

Whereas semi-supervised and unsupervised techniques are proposed when it is not possible to have an initial set of labeled documents/opinions to classify the rest of items

## 3.2 Lexicon-Based Approaches

Lexicon based method [20] uses sentiment dictionary with opinion words and match them with the data to determine polarity. They assigns sentiment scores to the opinion words describing how Positive, Negative and Objective the words contained in the dictionary are.

Lexicon-based approaches mainly rely on a sentiment lexicon, i.e., a collection of known and precompiled sentiment terms, phrases and even idioms, developed for traditional genres of communication, such as the Opinion Finder lexicon;

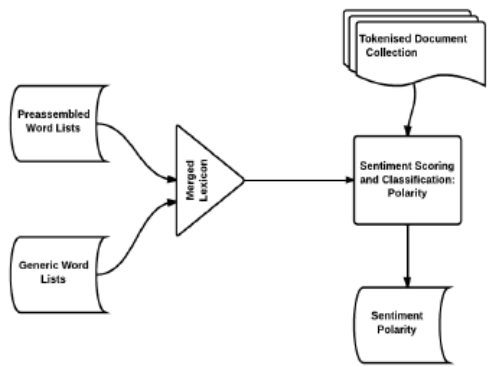

**Fig 3.Lexicon-Based Model**

There are Two sub classifications for this approach:

### 3.2.1.Dictionary-based:

It is based on the usage of terms (seeds) that are usually collected and annotated manually. This set grows by searching the synonyms and antonyms of a dictionary. An example of that dictionary is WordNet, which is used to develop a thesaurus called SentiWordNet.

**Drawback** : Can't deal with domain and context specific orientations.

### 3.2.2. Corpus-Based:

The corpus-based approach have objective of providing dictionaries related to a specific domain. These dictionaries are generated from a set of seed opinion terms that grows through the search of related words by means of the use of either statistical or semantic techniques.

- Methods based on statistics: Latent Semantic Analysis (LSA).
- Methods based on semantic such as the use of synonyms and antonyms or relationships from thesaurus like WordNet may also represent an interesting solution.

According to the performance measures like precision and recall, we provide a comparative study of existing techniques for opinion mining, including machine learning, lexicon-based approaches, cross domain and cross-lingual approaches, etc., as shown in Table 2.

**Table 2. Performance Comparison Of Sentiment Analysis Methods**

|  | Method | Data Set | Acc. | Author |
|---|---|---|---|---|
| Machine Learning | SVM | Movie reviews | 86.40% | Pang, Lee[23] |
|  | CoTraining SVM | Twitter | 82.52% | Liu[14] |
|  | Deep learning | Stanford Sentiment Treebank | 80.70% | Richard[18] |
| Lexical based | Corpus | Product reviews | 74.00% | Turkey |
|  | Dictionary | Amazon's Mechanical Turk | --- | Taboada[20] |
| Cross-lingual | Ensemble | Amazon | 81.00% | Wan,X[16] |
|  | Co-Train | Amazon, ITI68 | 81.30% | Wan,X.[16] |
|  | EWGA | IMDb movie review | >90% | Abbasi,A. |
|  | CLMM | MPQA,NTCIR,ISI | 83.02% | Mengi |
| Cross-domain | Active Learning | Book, DVD, Electronics, Kitchen | 80% (avg) | Li, S |
|  | Thesaurus |  |  | Bollegala[22] |
|  | SFA |  |  | Pan S J[15] |

## 4. SENTIMENT ANALYSIS TASKS

Sentiment analysis is a challenging interdisciplinary task which includes natural language processing, web mining and machine learning. It is a complex task and can be decomposed into following tasks, viz:

- Subjectivity Classification
- Sentiment Classification

- Complimentary Tasks

    o ObjectHolderExtraction
    o Object/Feature Extraction

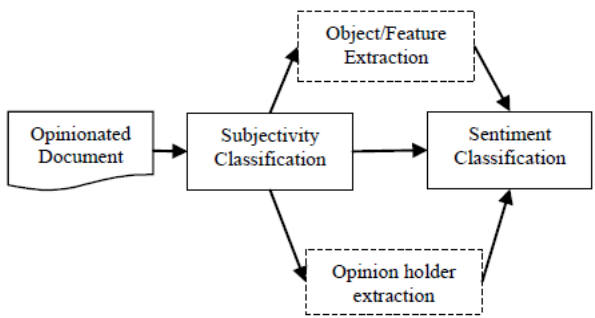

Fig.4 Sentiment Analysis Tasks

*A. Subjectivity classification*

Subjectivity classification is the task of classifying sentences as opinionated or not opinionated.

Let $S = \{s_1, \ldots, s_n\}$ be a set of sentences in document D. The problem of subjectivity classification is to identify sentences used to represent opinions and other forms of subjectivity(subjective sentences set $S_s$) from sentences used to objectively present factual information (objective sentences set $S_o$), where $S_s \cup S_o = S$.

*B. Sentiment Classification*

Once the task of finding whether a sentence is opinionated is done, we have to find the polarity of the sentence i.e., whether it expresses a positive or negative opinion. Sentiment classification can be a binary classification (positive or negative),multi-class classification(extremely negative, negative, neutral, positive or extremely positive),regression or ranking .

Depending upon the application of sentiment analysis, subtasks of opinion holder extraction and object feature extraction can be treated as optional.

*C. Complimentary Tasks*

- *OpinionHolder Extraction*

It is the discovery of opinion holders or sources. Detection of opinion holder is to recognize direct or indirect sources of opinion.

- *Object /Feature Extraction*

It is the discovery of the target entity.

## 5. LEVELS OF SENTIMENT ANALYSIS

Tasks described in the previous section can be done at several levels of granularity.

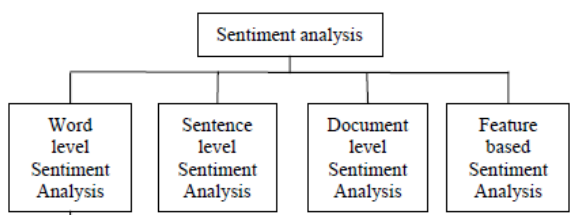

Fig.5 Levels Of Sentiment Analysis

### 5.1 Document level

It deals with tagging individual documents with their sentiment. In Document level the whole document is classify eitherinto positive or negative class.

*General Approach:*
Find the sentiment polarities of individual sentences or words and combine them together to find the polarity of the document.

*Other approaches:*

Complex linguistic phenomena like co-reference resolution, pragmatics, etc.

Various Tasks involved in this are:

- Task: Sentiment Classification of whole document

- Classes: Positive, negative and neutral

- Assumption :Each Document focuses on a single object (not true in discussion posts, blogs, etc.) and contain opinion from a single opinion holder

### 5.2 Sentence or phrase level

Sentence-level Sentiment Analysis deals with tagging individual sentences with their respective sentiment polarities. Sentence level sentiment classification classifies sentence into positive, negative or neutral class.

*General approach:*
find the sentiment orientation of individual words in the sentence/phrase and then to combine them to determine the sentiment of the whole sentence or phrase.

*Other approaches:*
consider discourse structure of the text

Various Tasks involved in this are:

- Task 1: Identifying Subjective/ Objective Sentences
    **Classes:** Objective and Subjective
- Task 2: Sentiment Classification of Sentences
    **Classes:** positive and negative
    **Assumption:** A sentence contains only one Opinion which may not always be true

### 5.3 Aspect level or Feature level

It deals with labeling each word with their sentiment and also identifying the entity towards which the sentiment is directed. Aspect or Feature level sentiment classification concerns with identifying and extracting product features from the source data. Techniques like dependency parser and discourse structures are used in this.

Various Tasks involved in this are:

- Task1: Identify and extract object features that have been commented on by an opinion holder (eg. A reviewer)

- Task2: Determining whether the opinions on features are negative, positive or neutral

- Task 3: Find feature synonyms

## 5.4 Word Level

Most recentworks have used the prior polarity of words and phrases for sentiment classification at sentence and document levels Word sentiment classification use mostly adjectives as features but adverbs,

The two methods of automatically annotating sentiment at the word level are:
(1) Dictionary-Based Approaches
(2) Corpus-Based Approaches.

## 6. EVALUATION OF SENTIMENT CLASSIFICATION

The performance of sentiment classification can be evaluated by using four indexes calculated as the following equations:

Accuracy = (TP+TN)/(TP+TN+FP+FN)

Precision = TP/(TP+FP)

Recall = TP/(TP+FN)

F1 = (2×Precision×Recall)/(Precision+Recall)

In which TP, FN, FP and TN refer respectively to the number of true positive instances, the number of false negativeinstances, the number of false positive instances and the number of true negative instances, as defined in the table 1.

**Table 3. Confusion Matrix**

|  | Predicted Positives | Predicted Negatives |
|---|---|---|
| Actual Positive | TP | FN |
| Actual Negative | FP | TN |

## 7. RESULTS AND DISCUSSION

We used the twitter dataset publicly made available by Stanford university. Analyses was done on this labeled datasets using various feature extraction technique. We used the framework where the preprocessor is applied to the raw sentences which make it more appropriate to understand. Further, the different machine learning techniques trains the dataset with feature vectors and then the semantic analysis offers a large set of synonyms and similarity which provides the polarity of the content.

**Dataset Description:**

| Train Data | 45000 |
|---|---|
| Negative | 23514 |
| Positive | 21486 |

| Test Data | 44832 |
|---|---|
| Negative | 22606 |
| Positive | 22226 |

**A. Baseline Algorithm:**

The baseline algorithm used is Naïve Bayeswithout preprocessed data and unigram model. Following table shows the accuracy obtained at different sizes for the baseline algorithm.

**Table 4. Accuracy of Baseline Algorithm**

| Dataset | Accuracy |
|---|---|
| 10 | 0.46475731620 |
| 50 | 0.533324411135 |
| 100 | 0.54744379015 |
| 500 | 0.612375089222 |
| 1000 | 0.652301927195 |
| 5000 | 0.697403640257 |
| 10000 | 0.712928265525 |
| 15000 | 0.717389364739 |
| 20000 | 0.722764989293 |
| 25000 | 0.729478943612 |
| 30000 | 0.729122055675 |
| 35000 | 0.73244557459 |
| 40000 | 0.733226266952 |
| 45000 | 0.736549785867 |

Following are the details on most informative features after the classifier is executed on train data.

sad = True          neg : pos   =   37.6 : 1.0

worst. = True       neg :pos    =   32.4 : 1.0

crying = True       neg : pos   =   24.7 : 1.0

fml = True          neg : pos   =   24.1 : 1.0

hurts = True        neg : pos   =   21.2 : 1.0

awful = True        neg : pos   =   21.1 : 1.0

ugh. = True         neg :pos    =   20.4 : 1.0

terrible = True     neg : pos   =   20.4 : 1.0
boo. = True         neg :pos    =   19.2 : 1.0
cancelled = True  neg : pos   =   19.2 : 1.0

**B. Naïve Bayes Algorithm:**

**Effect of Stopwords**
WhenNaiveBayes(Baseline)wasrun,itgaveanaccuracyof73.65percent,whichisconsideredasthebaselineresult.Thenextthingusedwasr e m o v a l          o f stopword.WhenstopwordswereremovedandNaiveBayeswasrun,itgaveanaccuracyof74.56percent.Following table shows the accuracy obtained at different sizes for the Naïve Bayes with stopwords removed and using preprocessed data and based on unigram model.

**Table 5. Accuracy of Naïve Bayes Algorithm (Stopword removal+unigram)**

| Dataset | Accuracy |
|---|---|
| 10 | 0.522305496074 |
| 50 | 0.583333333333 |
| 100 | 0.593839221984 |
| 500 | 0.649134546752 |
| 1000 | 0.673536759458 |
| 5000 | 0.7005710207 |
| 10000 | 0.717300142755 |
| 15000 | 0.725486259814 |
| 20000 | 0.731441827266 |
| 25000 | 0.734653818701 |
| 30000 | 0.738891862955 |
| 35000 | 0.740743219129 |
| 40000 | 0.742148465382 |
| 45000 | 0.745605817273 |

Most Informative FeaturesFor Naïve Bayes with stopwords removed and unigram model are,

bummed = True         neg : pos   =   34.8 : 1.0

```
disappointed = True   neg : pos   =   28.8 : 1.0
sad = True            neg : pos   =   27.6 : 1.0
awful = True          neg : pos   =   20.3 : 1.0
ugh = True            neg : pos   =   19.3 : 1.0
poor = True           neg : pos   =   19.3 : 1.0
sucks = True          neg : pos   =   18.7 : 1.0
upset = True          neg : pos   =   18.0 : 1.0
argh = True           neg : pos   =   17.3 : 1.0
battery = True        neg : pos   =   16.6 : 1.0
```

The results are slightly different; this was the case even with Linear SVC. This shows that stopwords really affect the predictions. An intuition to this can be obtained from the fact that given the short length of tweets, people generally use stopwords such as and, while, before, after and so on. Thus removal of stopwords makes a lot of difference to the accuracy.

**Effect of Bigram:**

Bigram uses a combination of two words as a feature. Bigram effectively captures some features in the data that unigram fails to capture. For example, words like 'not sad', 'not good' clearly say that the sentiment is negative. This effect can be clearly seen from the increase in accuracy from 74.56(Unigram) to 76.44 percent which is almost a 2% increase. Following table shows the accuracy obtained at different sizes for the Naïve Bayes algorithm with bigram model.

**Table 6. Accuracy of Naïve Bayes Algorithm (Stopword removal+Bigram)**

| Dataset | Accuracy |
|---|---|
| 10 | 0.544990185582 |
| 50 | 0.593593861527 |
| 100 | 0.591407922912 |
| 500 | 0.654956281228 |
| 1000 | 0.67193076374 |
| 5000 | 0.718214668094 |
| 10000 | 0.730973411849 |
| 15000 | 0.740609386153 |
| 20000 | 0.746431120628 |
| 25000 | 0.75073608137 |
| 30000 | 0.755041042113 |
| 35000 | 0.758453783012 |
| 40000 | 0.762892576731 |
| 45000 | 0.764476266952 |

The most informative features for Naive Bayes with Bigrams as features.

('so', 'sad') = True              neg :pos   =   55.2 : 1.0

sad. = True      neg :pos   =   44.2 : 1.0

bummed = True      neg : pos   =   33.8 : 1.0

horrible = True              neg : pos   =   32.0 : 1.0

('USERNAME', 'welcome') = True pos :neg   =   29.5 : 1.0

('welcome', 'to') = True         pos :neg   =   28.1 : 1.0

sad = True      neg : pos   =   27.5 : 1.0

('i', 'lost') = True         neg :pos   =   24.7 : 1.0

died = True      neg : pos   =   24.3 : 1.0

('miss', 'him') = True         neg :pos   =   24.1 : 1.0

**Effect of using Trigram:**

Running Naïve Bayes using Trigrams, bigrams and unigrams together gave an accuracy of 75.41 percent which is less than the accuracy obtained when Bigrams were used as a feature. Also this feature combination bloats up the feature space exponentially and the execution becomes extremely slow. Hence for further analysis, the trigrams are not considered as they do not have a notice able impact on the accuracy. Following table shows the accuracy obtained at different sizes for the Naïve Bayes algorithm with Trigram model.

**Table 7. Accuracy of Naïve Bayes Algorithm (Stopword removal+Trigram)**

| Dataset | Accuracy |
|---|---|
| 10 | 0.486995895789 |
| 50 | 0.528484118487 |
| 100 | 0.581571199143 |
| 500 | 0.634346002855 |
| 1000 | 0.654331727338 |
| 5000 | 0.703403818701 |
| 10000 | 0.721002855103 |
| 15000 | 0.731352605282 |
| 20000 | 0.737419700214 |
| 25000 | 0.742148465382 |
| 30000 | 0.74823786581 |
| 35000 | 0.748773197716 |
| 40000 | 0.753234296931 |
| 45000 | 0.754171127766 |

The most informative features for Naive Bayes with Trigrams as features.

('so', 'sad') = True           neg :pos   =   59.1 : 1.0

('lost', 'my') = True         neg :pos   =   38.9 : 1.0

('i', 'miss', 'my') = True     neg :pos   =   36.9 : 1.0

('going', 'to', 'miss') = True         neg :pos   =   28.5 : 1.0

('miss', 'him') = True         neg :pos   =   25.4 : 1.0

('happy', "mother's", 'day') = True  pos :neg   =   25.0 : 1.0

("can't", 'sleep') = True         neg :pos   =   21.5 : 1.0

('sad', 'that') = True         neg :pos   =   21.5 : 1.0

('miss', 'my') = True         neg :pos   =   21.4 : 1.0

('i', 'lost') = True         neg :pos   =   20.9 : 1.0

Following graph shows the summary of the results obtained by using different features and variation in the naïve bayes algorithm.

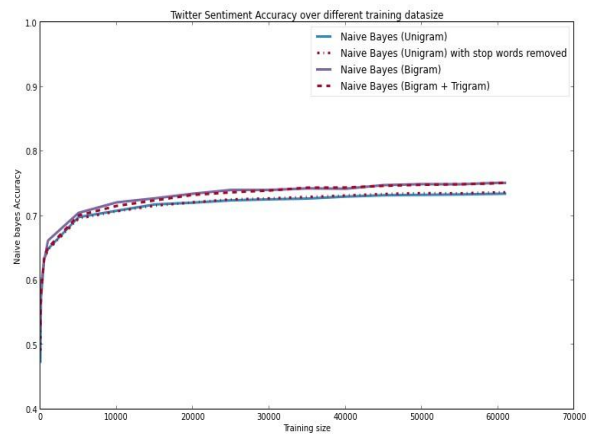

**Fig.6 Graph Representing Different results obtained for Naïve Bayes Algorithm.**

**Table 8. Accuracy of Naïve Bayes Algorithm**

| Algorithm | Accuracy |
|---|---|
| Naïve Bayes (unigram) | 74.56 |
| Naïve Bayes (bigram) | 76.44 |
| Naïve Bayes (trigram) | 75.41 |

### C. Support Vector Machine (SVM):
**Effect using unigram**
Following table shows the accuracy obtained at different sizes for the SVM algorithm with unigram model.

**Table 9. Accuracy of SVM Algorithm (Unigram)**

| Dataset | Accuracy |
|---|---|
| 10 | 0.525450571021 |
| 50 | 0.550521948608 |
| 100 | 0.569726980728 |
| 500 | 0.6261375803 |
| 1000 | 0.660421127766 |
| 5000 | 0.726222341185 |
| 10000 | 0.739806388294 |
| 15000 | 0.748973947181 |
| 20000 | 0.75426034975 |
| 25000 | 0.758096895075 |
| 30000 | 0.76130888651 |
| 35000 | 0.762847965739 |
| 40000 | 0.76556923626 |
| 45000 | 0.766862955032 |

**Effect using Bigram**
Following table shows the accuracy obtained at different sizes for the SVM algorithm with Bigram model.

**Table 10. Accuracy of SVM Algorithm (Bigram)**

| Dataset | Accuracy |
|---|---|
| 10 | 0.500223054961 |
| 50 | 0.574232690935 |
| 100 | 0.56437366167 |
| 500 | 0.632293897216 |
| 1000 | 0.657989828694 |
| 5000 | 0.725486259814 |
| 10000 | 0.746609564597 |
| 15000 | 0.756468593862 |
| 20000 | 0.761487330478 |
| 25000 | 0.767375981442 |
| 30000 | 0.771011777302 |
| 35000 | 0.77210474661 |
| 40000 | 0.775941291934 |
| 45000 | 0.777324232691 |

**Table 10. Summary for Accuracy of SVM Algorithm**

| Algorithm | Accuracy |
|---|---|
| SVM with unigram | 76.68 |
| SVM with bigram | 77.73 |

### D. Maximum Entropy
In Maximum Entropy Classifier, no assumptions are taken regarding the relationship between features.we obtained an accuracy of 74.93 percent with unigram model

With all the features considered, the results show that SVM outperforms NaiveBayes and maximum entropy as well in all cases .In particular, the feature combination of Slang stopwords removal and Bigram gives the maximum accuracy of 77.73 with SVM. Maximum Entropy model gives an accuracy consistently in-between NaiveBayes and SVM. Also it runs iteratively and takes a large amount of time to run. Hence MaxEnt was not used for all the feature combinations.

**Table 11. Summary Of Results For Unigram**

| Method | Accuracy(Unigram) |
|---|---|
| Baseline | 73.65 |
| Naïve Bayes | 74.56 |
| SVM | 76.68 |
| Maximum Entropy | 74.93 |

As the table shows, when the processing, analysis was done on the bigger dataset, the accuracy scaled upto a great extent. NaiveBayes baseline scaled upto 76.44 and SVM scaled upto 77.73percent.The best result tested thus far, was obtained when SVM was used on a feature set of a combination of Unigram, Bigram with stopwords removal, gave an accuracy of 77.73. MaxEnt also performed well and gave an accuracy of 74.93when stopwords was removed.

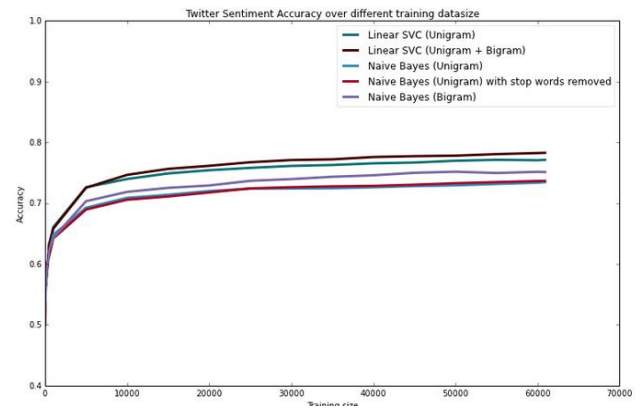

**Fig.7 Graph Representing Different results obtained for Naïve Bayes Algorithm And Linear SVC (SVM).**

## 8. CHALLENGES IN SENTIMENT ANALYSIS

Sentiment Analysis is a very challenging task. Following are some of the challenges[13] faced in Sentiment Analysis of Twitter.

**1. Identifying subjective parts of text:**
Subjective parts represent sentiment-bearing content. The same word can be treated as subjective in one case, or an objective in some other. This makes it difficult to identify the subjective portions of text.

For **example**:
1. The language of the Mr.Dennis was very crude.
2. Crude oil is obtained by extraction from the sea beds.
The word 'crude' is used as an opinion in first example, while it is completely objective inthe second example.

**2. Domain dependence[24]:**
The same sentence or phrase can have different meanings in different domains. For Example, the word 'unpredictable' is positive in the domain of movies, dramas ,etc, but if the same word is used in the context of a vehicle's steering, then it has a negative opinion.

**3. Sarcasm Detection:**
Sarcastic sentences express negative opinion about a target using positive words in unique way..
*Example*:

"Nice perfume. You must shower in it."
The sentence contains only positive words but actually it expresses a negative sentiment.

**4. Thwarted expressions:**
There are some sentences in which only some part of text determines the overall polarity of the document.

*Example:*

"This Movie should be amazing. It sounds like a great plot, the popular actors , and the supporting cast is talented as well. "

In this case,a simple bag-of-words approaches will term it as positive sentiment, but the ultimate sentiment is negative.

**5. Explicit Negation of sentiment:**
Sentiment can be negated in many ways as opposed to using simple no, not, never, etc. It is difficult to identify such negations .

*Example:*

"It avoids all suspense and predictability found in Hollywood movies."

Here the words suspense and predictable bear a negative sentiment, the usage of 'avoids' negatestheir respective sentiments.

**6. Order dependence:**
Discourse Structure analysis is essential for Sentiment Analysis/Opinion Mining.
*Example:*
A is better than B, conveys the exact opposite opinion from, B is better than A.

**7. Entity Recognition:**
There is a need to separate out the text about a specific entity and then analyze  sentiment towards it.

*Example***:**

"I hate Microsoft, but I like Linux".

A simple bag-of-words approach will label it as neutral, however, it carries a specific sentiment for both the entities present in the statement.

**8. Building a classifier for subjective vs. objective tweets.**
Current research work focuses mostly on classifying positive vs. negative correctly. There is need to look at classifying tweets with sentiment vs. no sentiment closely.

**9. Handling comparisons.**
Bag of words model doesn't handle comparisons very well.
*Example*:

"IIT's are better than most of the private colleges", the tweet would  be considered positive  for  both  IIT's  and  private colleges using bag of words model because it doesn't take into account the relation towards "better".

**10. Applying sentiment analysis to Facebook messages.**
There has been less work on sentiment analysis on Facebook data mainly due to various restrictions by Facebook graph api and security policies in accessing data.

**11. Internationalization [16,17].**
Current Research work focus mainly on English content, but Twitter has many varied users from across.

## 9. APPLICATIONS OF SENTIMENT ANALYSIS
Sentiment Analysis has many applications in various Fields.

*1.Applications that use  Reviewsfrom Websites:*

Today Internet has a large collection of reviews and feedbacks on almost everything. This includes product reviews, feedbacks on political issues, comments about services, etc. Thus there is a need for a sentiment analysis system that can extract sentiments about a particular product or services. It will help us to automate in provision of feedback or rating for the given product, item, etc. This would serve the needs of both the users and the vendors.

*2. Applications as a Sub-component Technology*

A sentiment predictor system can be helpful in recommender systems as well. The recommender system will not recommend items that receive a lot of negative feedback or fewer ratings.

In online communication, we come across abusive language and other negative elements. These can  be detected simply by identifying a highly negative sentiment and correspondingly taking action against it.

*3. Applications in Business Intelligence*

It has been observed that people nowadays tend to look upon reviews of products which are available online before they buy them. And for many businesses, the online opinion decides the success or failure of their product. Thus, Sentiment Analysis plays an important role in businesses. Businesses also wish to extract sentiment from the online reviews in order to improve their products and in turn their reputation and help in customer satisfaction .

*4. Applications across Domains:*

Recentresearches in sociology and other fields like medical, sports have also been benefitted by Sentiment Analysis that show trends in human emotions especially on social media.

*5. Applications In Smart Homes*

Smart homes are supposed to be the technology of the future. In future entire homes would be networked and people would be able to control any part of the home using a tablet device. Recently there has been lot of research going on Internet of Things(IoT). Sentiment Analysis would also find its way in IoT. Like for example, based on the current sentiment or emotion of the user, the home could alter its ambiance to create a soothing and peaceful environment.

Sentiment Analysis can also be used in trend prediction. By tracking public views, important data regarding sales trends and customer satisfaction can be extracted.

## 10. CONCLUSION
In this paper, we provide a survey and comparative study of existing techniques for opinion mining including machine learning and lexicon-based approaches, together with cross domain and cross-lingual methods and some evaluation metrics. Research results show that machine learning methods, such as SVM and naive Bayes have the highest accuracy and can be regarded as the baseline learning methods, while lexicon-based methods are very effective in some cases, which require few effort in human-labeled document .We also studied the effects of various features on classifier. We can conclude that more the cleaner data, more accurate results can be obtained. Use of bigram model provides better sentiment accuracy as compared to other models. We can focus on the study of combining machine learning method into opinion lexicon method in order to improve the accuracy of sentiment classification and adaptive capacity to variety of domains and different languages.